# Exploiting Feature and Class Relationships in Video Categorization with Regularized Deep Neural Networks

Yu-Gang Jiang, Zuxuan Wu, Jun Wang, Xiangyang Xue, and Shih-Fu Chang, *Fellow, IEEE*

**Abstract**—In this paper, we study the challenging problem of categorizing videos according to high-level semantics such as the existence of a particular human action or a complex event. Although extensive efforts have been devoted in recent years, most existing works combined multiple video features using simple fusion strategies and neglected the utilization of inter-class semantic relationships. This paper proposes a novel unified framework that jointly exploits the feature relationships and the class relationships for improved categorization performance. Specifically, these two types of relationships are estimated and utilized by imposing regularizations in the learning process of a deep neural network (DNN). Through arming the DNN with better capability of harnessing both the feature and the class relationships, the proposed regularized DNN (rDNN) is more suitable for modeling video semantics. We show that rDNN produces better performance over several state-of-the-art approaches. Competitive results are reported on the well-known Hollywood2 and Columbia Consumer Video benchmarks. In addition, to stimulate future research on large scale video categorization, we collect and release a new benchmark dataset, called FCVID, which contains 91,223 Internet videos and 239 manually annotated categories.

**Index Terms**—Video categorization, deep neural networks, regularization, feature fusion, class relationships, benchmark dataset

✦

## 1 INTRODUCTION

VIDEOS carry very rich and complex semantics, and are intrinsically multimodal. Techniques for recognizing high-level semantics in diverse unconstrained videos can be deployed in many applications, such as Internet video search. However, it is well-known that semantic recognition or categorization of videos is an extremely challenging task due to the semantic gap between computable low-level video features and the complex high-level semantics. While significant progress has been achieved in recent years, most state-of-the-art solutions rely on a large set of features to recognize a class-of-interest. In order to derive a robust fused representation that bridges the semantic gap, the fusion process of multiple features is usually expected to learn the cross-feature correlations, such as the relationship of HOG and HOF features that model visual information and their complements to acoustic descriptors. In addition to the feature relationships, there are also certain correlations among multiple high-level semantic categories: knowing the presence of one category may provide useful clues for recognizing other related categories. For example, a high score of a video clip containing "running" ("diving") will increase (decrease) the confidence of the video containing "soccer". Although there exist several works investigating multi-feature fusion or exploiting the inter-class relationships, as will be discussed in the next section, they mostly address the two problems separately.

Motivated by the limitations of the existing techniques and the increasing popularity of using Deep Neural Networks (DNN) for visual categorization, in this paper we propose a novel unified framework that jointly learns the feature relationships and the class relationships using a DNN. Video categorization can also be carried out within the same network utilizing the learned relationships.

Fig. 1 gives an overview of the proposed approach. We first extract several popular video features, including the popular frame-based features computed by the convolutional neural networks (CNN) [1], trajectory-based motion descriptors and audio descriptors. The features are then used as the inputs of a DNN, where the first two layers are input and feature transformation layers, respectively. The third layer is called fusion layer, where we impose regularization on the network weights to identify and utilize the feature relationships. Specifically, the regularization terms are selected based on two natural properties of the inter-feature relationships: correlation and diversity. The former means that different features may share some common patterns in a middle level representation lying between the original features and the high-level semantics (i.e., the transformed features after the second layer). The latter emphasizes the

- *Y.-G. Jiang, Z. Wu, and X. Xue are with the School of Computer Science, Fudan University, Shanghai 200433, China.*
  *E-mail: {ygj, zxwu, xyxue}@fudan.edu.cn.*
- *J. Wang is with the School of Computer Science and Software Engineering, East China Normal University, Shanghai 200062, China.*
  *E-mail: wongjun@gmail.com.*
- *S.-F. Chang is with the Department of Electrical Engineering, Columbia University, New York, NY 10027. E-mail: sfchang@ee.columbia.edu.*

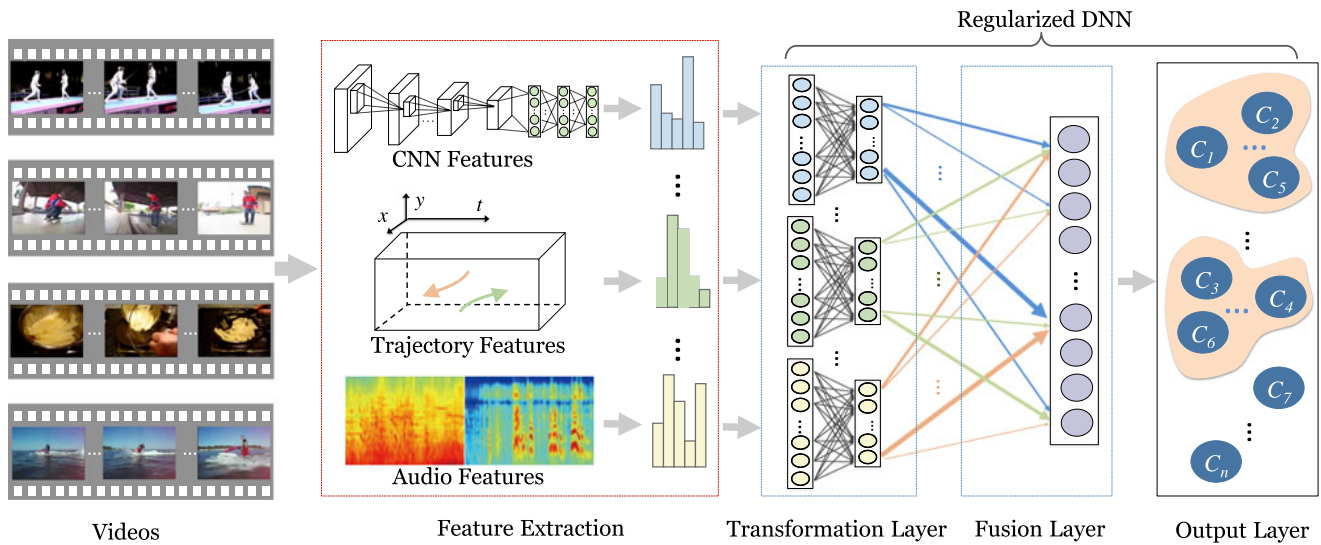

Fig. 1. Illustration of the proposed rDNN framework for video categorization. See texts for more explanations.

unique characteristics of different features, which are the complementary information that is likely to be useful for identifying video semantics. Through modeling both properties using a feature correlation matrix, we impose a trace-norm regularization over the network weights to reveal the hidden correlations and diversity of the features.

In order to discover and utilize the inter-class relationships, we impose similar regularizations on the weights of the final output layer. This helps to identify the grouping structures of video classes and the outlier classes. Semantic classes within the same group share commonalities that can be utilized as knowledge sharing for improved categorization performance, while the outlier classes should be excluded from "negative" knowledge sharing. By executing regularized learning of the two kinds of relationships within the same DNN, we arrive at a unified framework, namely *regularized DNN* (rDNN), which can be easily implemented.

Although the framework shown in Fig. 1 is built on the static CNN feature and a few typical hand-crafted video features, it is feasible to use our approach for fusing any features. We also realize that, in the image categorization domain, the CNN features are dominating state-of-the-art approaches. The reasons of considering both the CNN feature and the hand-crafted features in this work are two-folds. First, the hand-crafted features have been widely used for video categorization and remain the key components of many systems that generated recent state-of-the-art results on tasks like human action recognition [2] and complex event recognition [3], [4]. By using these features it is easy to make comparisons with the traditional approaches. Second, so far, very few existing works on neural networks based video feature extraction have demonstrated significantly better results than the traditional hand-crafted features. Some end-to-end learning methods only showed lower or similar results [5], [6], and, similar to this paper, a recent work [7] reported better results by combining deep features with hand-crafted features. Therefore, this paper considers both the deeply learned and the hand-crafted features, and focuses on the tasks of feature fusion and semantic categorization.

The main contribution of this paper is the proposal of the rDNN. To the best of our knowledge, rDNN is the first attempt to exploit both the feature and the class relationships in the DNN pipeline for video categorization. Our formulation is designed to model the complex relationships such as feature/class correlation and diversity. It is easy to implement and can be efficiently executed using a GPU. In addition, we introduce and release a new benchmark dataset, called Fudan-Columbia Video Dataset (FCVID). FCVID contains 91,223 YouTube videos and 239 manually annotated categories. It is currently one of the largest manually annotated datasets of Internet videos. Compared with some recently released video benchmarks, FCVID covers a wide range of categories popularly seen in Internet user-shared videos, including events, scenes and objects. For example, the new EventNet [8] consists of only events and its labels are noisy as it is not manually labeled; the Sports-1M dataset [5] focuses only on sports and is also not manually labeled; the ActivityNet [9] focuses on human actions; and, the MPII Human Pose dataset was mainly designed for recognizing human poses [10]. We evaluate rDNN using our new FCVID dataset, and hope that its public release could stimulate future research around this challenging problem.

This work is based upon a conference publication [11] with the following major extensions. First, a more comprehensive survey of the state of the arts on video categorization is included in the next section. Second, we provide proofs and more discussions on how the learned relationships could help improve the recognition performance. Third, we leverage more powerful feature representations (i.e., the CNN features) to evaluate the generalization ability of the framework, and implement several additional alternative methods to fully justify the effectiveness of our approach. Finally, we introduce a new dataset that is much larger than those popularly used in recent literature. The rest of this paper is organized as follows. Section 2 discusses related works, where we mainly focus on the existing works exploiting feature or class relationships. Section 3 elaborates the proposed rDNN approach. Extensive experimental results are discussed in Section 4, where we also briefly introduce the new FCVID dataset. Finally, Section 5 concludes this paper.

## 2 RELATED WORK

Video categorization has received significant research attention. Most approaches followed a very standard pipeline,

where various features are first extracted and then used as inputs of classifiers. Many works have focused on the design of novel features, such as the biologically inspired pipeline [12], Spatial-Temporal Interest Points (STIP) [13], trajectory-based descriptors [2], audio clues [14], and the Convolutional Neural Networks based features [1], [5], [6], [15].

In contrast to the variety of video features, Support Vector Machines (SVM) have been the dominate classifier option for over a decade. Recently, with the increasing popularity of the deep learning based approaches, neural networks have also been adopted for video classification [5], [6], [15]. Among them, probably the most well-known deep learning based video categorization result is probably from Simonyan and Zisserman [6], who used a two-stream CNN approach to extract features from static frames and motion optical flow respectively. The features were classified separately and the predictions were then simply fused with fixed weights. Using this pipeline, they reported similar performance to the improved dense trajectories [2], one of the best hand-crafted feature-based approaches. More recently, in addition to the CNN, researchers also adopted recurrent neural networks (RNN) to model the long-term temporal information in videos [16], [17], [18] and reported promising results.

Besides accuracy, efficiency is another important factor that should be considered in the design of a modern video classification system. Several recent studies investigated this issue by proposing efficient classification methods [19], [20] or parallel computing strategies [21], [22].

In the following we primarily discuss works on multi-feature fusion and/or exploiting class relationships, which are more closely related to this work.

### 2.1 Exploiting Feature Relationships

In most state-of-the-art video categorization systems, two feature fusion strategies were popularly adopted, i.e., the early fusion and the late fusion. Early fusion relies on the assumption that multiple features are explicitly complementary to each other, however this assumption does not always hold in the complex video data. Late fusion trains models separately and then combines prediction scores. This method cannot explore feature relationships in the categorization process as the features are processed separately. In contrast, this work derives a fused representation by explicitly regularizing the fusion process, and the fusion process and classification are conducted simultaneously under a unified objective. In other words, our approach intends to learn what features are correlated and what are unique clues that exist only in one input feature. These learned information is used in generating the fused representation.

In both early and late fusion, fusion weights are needed to weigh the importance of each individual feature dimension, which can be set as equal values (a.k.a. average fusion) or learned based on cross validation. In several recent works, multiple kernel learning (MKL) [23] was adopted to estimate the fusion weights [24], [25]. MKL was reported to produce better performance in some cases, but the gain was also often observed to be insignificant [26].

Several more advanced feature fusion approaches were proposed. In [27], Ye et al. proposed an optimization framework, called robust late fusion, which uses a shared low-rank matrix to remove noises in certain feature modalities. In [28], Jiang et al. focused on exploiting the correlations between audio and visual features. They proposed to generate an audio-visual joint codebook by discovering the correlations of the two features for video classification. Jhuo et al. [29] followed a similar framework, and improved the speed of training the audio-visual codebook by replacing the segmentation-based region features with standard local features.

With the growing popularity of the DNN, a few recent studies focused on combining multiple features in neural networks, which are closely related to this work. A deep denoised auto-encoder was employed in [30] to learn a shared representation based on mutimodal inputs. Similarly, a deep Boltzmann machine was utilized in [31] to fuse visual and textual features. Very recently, Kihyuk et al. [32] proposed to learn a good shared representation by minimizing variation of information, so that missing input modality can be better predicted based on the available information. They showed that this method outperforms [31] on several image classification benchmarks. Different from [30], [31] that fused the features in a "free" way, in this paper we propose *regularized fusion* of multiple features, which is intuitively reasonable and empirically effective. Compared with [32], our objective is to learn and use dimension-wise feature relationships. Minimizing the variation of information in [32] might be more suitable for images, but for videos, different modalities (e.g., audio and visual) may represent very distinctive information and simply minimizing their variation is not a good strategy to exploit the complementary information.

### 2.2 Exploiting Class Relationships

Many researchers have investigated class relationships, commonly termed context, to improve classification performance. In [33], Torralba et al. discussed the importance of context in the task of object detection in images. In [34], [35], the class co-occurrence context was utilized to improve object recognition accuracy. In the context of video analysis, Naphade and Huang [36] proposed to utilize a probabilistic graphical network to model the co-occurrence of semantic concepts for video indexing and retrieval. Jiang et al. [37] proposed a semantic diffusion algorithm to harness the class relationships. Weng et al. [38] proposed a similar domain-adaptive method that not only used the class relationships, but also explored temporal context information of broadcast news videos. Recently, Deng et al. [39] proposed Hierarchy and Exclusion (HEX) graphs, which can capture not only the co-occurrence class relationships, but also mutual exclusion and subsumption. Another two recent works [40], [41] utilized the co-occurrence statistics to help video classification, where the co-occurrence of classes was used more as a semantic feature representation.

Most of these approaches, however, rely on the co-occurrence statistics of the video classes, and thus cannot be used in the cases where the classes share certain commonalities but do not explicitly co-occur in the same video. By injecting a class relationship matrix into the learning process, our approach can automatically learn and utilize such commonalities using a regularized DNN with a unified objective, as will be elaborated in the following section.

Our formulation is partly inspired by recent research on Multiple Task Learning (MTL) [42], [43]. MTL trains multiple class models simultaneously and boosts the performance of a task (a classifier model) by seeking help from other related tasks. MTL has demonstrated good results in many applications, such as disease prediction [44], [45] and financial stock selection [46]. Sharing certain commonalities among multiple tasks is the key idea of MTL and several algorithms have been proposed with regularizations on the shared patterns across tasks [47], [48], [49]. These works exploited the class relationships in classification or regression problems using the conventional learning approaches, but never injected such regularizations into the DNN.

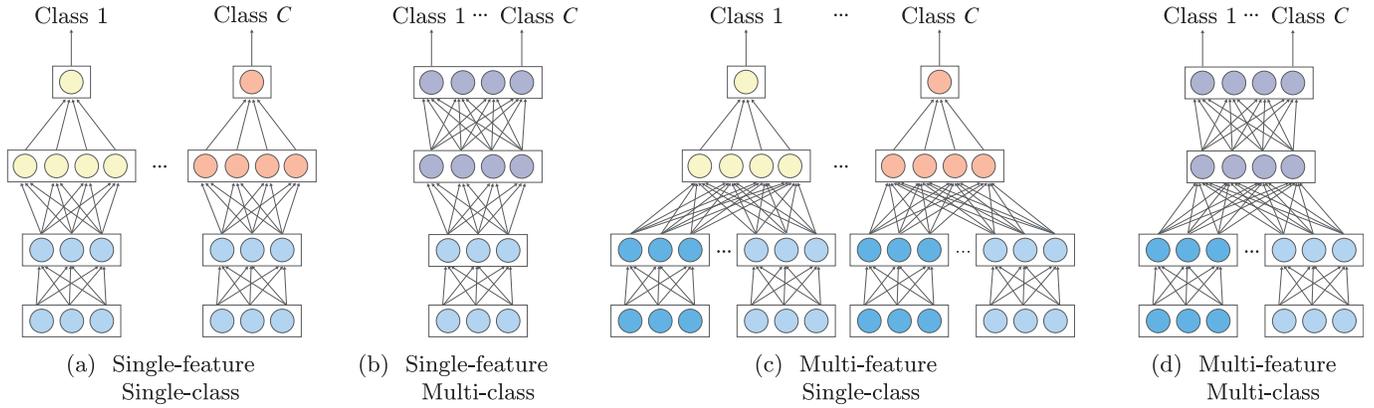

Fig. 2. Popular neural network structures: (a) Is the standard one-vs-all training scheme using a single type of feature; (b) is the popular structure used in multi-class learning with a single type of feature; (c) is the one-vs-all training scheme using multiple types of features; and (d) processes multiple features separately and then performs fusion using a middle layer [31].

In fact, neural network is one of the earliest MTL models [50]. See Fig. 2b for a standard network structure. In the network, each unit of the output layer refers to a task (class) and neurons of the hidden layers can be viewed as the shared common features. In this paper, we show that, by imposing explicit forms of regularizations, the class relationships can be better exploited for improved video categorization performance.

## 3 REGULARIZED DNN

### 3.1 Notations and Settings

We have a training set with a total of $N$ video samples, which are associated with $C$ semantic classes. Since a video sample may have $M$ types of feature representations (e.g., multiple visual and audio clues), we can use an $(M+1)$-tuple to represent each video as

$$(\mathbf{x}_n^1, \ldots, \mathbf{x}_n^m, \ldots, \mathbf{x}_n^M, \mathbf{y}_n), n = 1, \ldots, N.$$

Here $\mathbf{x}_n^m$ represents the $m$th feature of the $n$th video sample, and $\mathbf{y}_n = [y_{n1}, \ldots y_{nc}, \ldots y_{nC}]^\top \in \mathbb{B}^C$ is the associated semantic label for the $n$th sample. If the $n$th sample belongs to the $c$th semantic class, the $c$th element is set as $y_{nc} = 1$, otherwise $y_{nc} = 0$. The objective for video classification under the above setting is to train prediction models that can categorize new test videos into the $C$ semantic classes.

Simply, one can independently train one classifier for each semantic class, where different features can be combined using either the early or the late fusion scheme. Instead, here we propose a DNN framework with structure regularization to perform video classification. In particular, this regularized DNN carries out *feature fusion* with an additional layer, namely fusion layer, to exploit the correlation and diversity of multiple features, as illustrated in Fig. 1. Furthermore, we impose additional regularization on the prediction layer to enforce *knowledge sharing* across different semantic classes. With such a regularized DNN framework, we are able to explicitly explore both types of relationships in a uniform learning process. To address the details of the proposed regularized DNN, below we first introduce the background of training standard DNNs with a single type of feature. After that, we elaborate our formulation and explain why our proposed approach can realize the aforementioned goals.

### 3.2 DNN Learning with a Single Type of Feature

Inspired by the biological neural systems, DNN uses a large number of interconnected neurons and construct complex computational models to mimic the information processing in neural systems. Through cascading the neurons in multiple layers, DNN exhibits strong non-linear abstraction capacity and is able to learn arbitrary mapping from inputs to outputs as long as being given sufficient training data.

Given a DNN with $L$ layers, we denote $\mathbf{a}_{l-1}$ and $\mathbf{a}_l$ as the input and the output of the $l$th layer, $l = 1, \ldots, L$, while $\mathbf{W}_l$ and $\mathbf{b}_l$ refer to the weight matrix and the bias vector of the $l$th layer, respectively. With only a single type of feature, the transition function from the $(l-1)$th layer to the $l$th layer can be written as

$$\mathbf{a}_l = \begin{cases} \sigma(\mathbf{W}_{l-1}\mathbf{a}_{l-1} + \mathbf{b}_{l-1}) & l > 1; \\ \mathbf{x} & l = 1, \end{cases} \quad (1)$$

where the nonlinear sigmoid function $\sigma(\cdot)$ is defined as

$$\sigma(\mathbf{x}) = \frac{1}{1 + e^{-\mathbf{x}}}.$$

For simplicity, we can absorb $\mathbf{b}_{l-1}$ into the weights coefficient $\mathbf{W}_{l-1}$ by adding an additional dimension to the feature vectors with a constant value one. Figs. 2a and 2b show two types of four-layered neural networks using a single feature as the input to classify data samples into $C$ semantic classes.

Typically, one can minimize the following cost function to derive the optimal weights for each layer in the network

$$\min_{\mathbf{W}} \quad \sum_{i=1}^{N} \ell(f(\mathbf{x}_i), \mathbf{y}_i) + \frac{\lambda_1}{2} \sum_{l=1}^{L-1} \|\mathbf{W}_l\|_F^2. \quad (2)$$

The first part in the above cost function measures the empirical loss on the training data, which summarizes the discrepancy between the outputs of the network $\hat{\mathbf{y}}_i = \mathbf{a}_L = f(\mathbf{x}_i)$ and the ground-truth labels $\mathbf{y}_i$. The second part is a regularization term preventing overfitting.

### 3.3 Regularization for Feature Fusion

The DNN using a single type of feature is effective in some cases. However, for data with a variety of representations like videos, the semantics could be carried by different feature representations. Motivated by the multisensory integration process of primary neurons in biological systems [51], [52], we extend the basic DNN with structure regularization on an additional fusion layer to accommodate the *deep fusion* process of multiple types of features. As demonstrated in Fig. 1, the fusion layer absorbs all the outputs from the transformation layer to generate an integrated representation as the input for the classification layer. Accordingly, the transition equation for this fusion layer can be written as the following:

$$\mathbf{a}_F = \sigma \left( \sum_{m=1}^{M} \mathbf{W}_E^m \mathbf{a}_E^m + \mathbf{b}_E \right). \quad (3)$$

We denote $E$ as the index of the last layer of feature transformation and $F$ as the index of the fusion layer (i.e., $F = E + 1$). Hence, $\mathbf{a}_E^m$ represents the extracted mid-level representation for the $m$th feature. From the above transition equation, the mid-level representation is first linearly transformed by the weight matrix $\mathbf{W}_E^m$ and then non-linearly mapped to generate the fused representation $\mathbf{a}_F$ using a sigmoid function.

Note that the weights of the fusion layer, $\mathbf{W}_E^1, \ldots, \mathbf{W}_E^M$, transform all the available features into a shared representation. Here the weight matrices are first vectorized into $P$ dimensional vectors separately with $P = |\mathbf{a}_E^m| \cdot |\mathbf{a}_F|$ being the product of the $\mathbf{a}_E^m$'s ($m = 1, \ldots, M$) dimension and the $\mathbf{a}_F$'s dimension. To simplify the formulation, we assume the extracted features $\mathbf{a}_E^m$ are of the same dimension. Then all the coefficient vectors are stacked into a matrix $\mathbf{W}_E \in \mathbb{R}^{P \times M}$. Each column of $\mathbf{W}_E$ corresponds to the weights of a single feature with the element $\mathbf{W}_E(i, j)$ given as

$$\mathbf{W}_E(i, j) = \mathbf{W}_E^i(j), \quad i = 1, \ldots, M, \quad j = 1, \ldots, P.$$

In order to perform *feature fusion* by exploring correlations and diversities simultaneously, we formulate the following regularized optimization problem to learn the weights of the DNN

$$\begin{aligned}
\min_{\mathbf{W}, \Psi} \quad & \mathcal{L} + \frac{\lambda_1}{2} \left( \sum_{l=1}^{E} \sum_{m=1}^{M} \|\mathbf{W}_l^m\|_F^2 + \sum_{l=F}^{L-1} \|\mathbf{W}_l\|_F^2 \right) \\
& + \frac{\lambda_2}{2} \mathrm{tr}(\mathbf{W}_E \Psi^{-1} \mathbf{W}_E^\top) \\
\text{s.t.} \quad & \Psi \succeq 0,
\end{aligned} \quad (4)$$

where $\mathcal{L} = \sum_{i=1}^{N} \ell(\hat{\mathbf{y}}_i, \mathbf{y}_i)$ is the empirical loss term. Different from the standard single feature based neural network (Equation (2)), we include one additional regularization term in the above cost function with one more variable $\Psi \in \mathbb{R}^{M \times M}$ to model the inter-feature correlation.

Note that $\Psi$ is a symmetric and positive semidefinite matrix and the last regularization term with the trace norm can help utilize the inter-feature relationship. Similar formulations were often used in multiple task learning [43], [53], where task relationships are explored to improve the learning performance. Intuitively, the goal is to ensure that the weight vectors of correlated feature dimensions should contain similar values so that the correlated feature dimensions can contribute similarly to the fused representation. On the one hand, if a non-diagonal entry of $\Psi$ is large, updating $\mathbf{W}_E$ by minimizing the trace norm ensures that the weights of the corresponding feature dimensions are similar. On the other hand, if $\mathbf{W}_E$ is fixed, minimizing the trace norm can help learn a $\Psi$ with entries more consistent with the network weights. Please see Equations (9), (10), and (11) for a proof on the relationships between $\mathbf{W}_E$ and $\Psi$. In the optimization stage, we adopt an alternative minimization strategy to learn $\mathbf{W}_E$ and $\Psi$ together, as will be explained in Section 3.6. The coefficients $\lambda_1$ and $\lambda_2$ balance the contributions from different regularization terms.

### 3.4 Regularization for Class Knowledge Sharing

As discussed earlier, one can simply adopt the one-versus-all strategy to independently train $C$ classifiers for categorizing video semantics. As illustrated in Figs. 2a and 2c, this one-vs-all training scheme with a total of $C$ four-layered neural networks can be applied for both single-feature and multi-feature settings. To train a total of $C$ neural networks separately, a sufficient amount of positive training samples are desired for each video category. In addition, the independent training process completely neglects the knowledge sharing among different semantic categories. However, video semantics often share some *commonality* due to the strong correlations between different semantic categories, which have been observed in previous studies [37], [54], [55]. Therefore, it is critical to explore such a commonality by simultaneously learning multiple video semantics, which can lead to better learning performance [55]. Generally, the commonality among multiple classes is represented by the parameter sharing among different prediction models [56], [57]. In addition, it is fairly natural for DNN to perform simultaneous multi-class training. For example, as seen in Fig. 2b, by adopting a set of $C$ units in the output layer, a single-feature based DNN can be easily extended to multi-class problems.

Motivated by the regularization methods adopted for MTL [56], [57], here we present a regularized DNN that aims at training multiple classifiers simultaneously with deeper exploitation of the class relationships. To enforce class knowledge sharing, we employ the following optimization problem as our learning objective

$$\begin{aligned}
\min_{\mathbf{W}, \Omega} \quad & \sum_{i=1}^{N} \ell(f(\mathbf{x}_i), \mathbf{y}_i) + \frac{\lambda_1}{2} \sum_{l=1}^{L-1} \|\mathbf{W}_l\|_F^2 \\
& + \lambda_2 \mathrm{tr}(\mathbf{W}_{L-1} \Omega^{-1} \mathbf{W}_{L-1}^\top). \\
\text{s.t.} \quad & \Omega \succeq 0.
\end{aligned} \quad (5)$$

Although some previous MTL works explore similar regularization in the learning objective, they often assume that the class relationships are explicitly given and are ready for use as prior knowledge [43], [57]. In our formulation, we tend to learn the prediction model as well as the class relationships. In particular, we adopt a convex formulation by imposing a trace norm regularization term over the coefficients of the output layer $\mathbf{W}_{L-1}$ with the class relationships augmented as a matrix variable $\Omega \in \mathbb{R}^{C \times C}$. The constraint $\Omega \succeq 0$ indicates that the class relationship matrix is positive semidefinite since it can be viewed as the similarity measure of the semantic classes. The form of this regularization term is the same with the feature regularization in Equation (4), and minimizing it ensures the consistency between weight correlations in $\mathbf{W}_{L-1}$ and the non-diagonal values in $\Omega$. The coefficients $\lambda_1$ and $\lambda_2$ are regularization parameters that balance the contributions from different terms.

### 3.5 Final Objective of rDNN

Considering both objectives of feature fusion and class knowledge sharing, we now present a unified DNN formulation that is able to explore both the feature and the class relationships. In this joint framework, one additional layer is employed to fuse multiple features, where the objective is to bridge the gap between low-level features and the high-level video semantics. Then another layer of neurons is stacked over the fusion layer to generate the predictions, where we impose the trace norm regularization over the prediction models to encourage knowledge sharing across different semantic categories. To build such a rDNN, we incorporate both the feature regularization in Equation (4) and the class regularization in Equation (5) to form the following objective

$$\begin{aligned}
\min_{\mathbf{W},\Psi,\Omega} \quad & \mathcal{L} + \frac{\lambda_1}{2}\left(\sum_{l=1}^{E}\sum_{m=1}^{M}\|\mathbf{W}_l^m\|_F^2 + \sum_{l=F}^{L-1}\|\mathbf{W}_l\|_F^2\right) \\
& + \frac{\lambda_2}{2}\text{tr}(\mathbf{W}_E\Psi^{-1}\mathbf{W}_E^\top) \\
& + \frac{\lambda_3}{2}\text{tr}(\mathbf{W}_{L-1}\Omega^{-1}\mathbf{W}_{L-1}^\top), \\
\text{s.t.} \quad & \Psi \succeq 0 \quad \text{tr}(\Psi) = 1, \\
& \Omega \succeq 0 \quad \text{tr}(\Omega) = 1,
\end{aligned} \quad (6)$$

where $\lambda_1, \lambda_2$, and $\lambda_3$ are regularization parameters. In the above formulation, two trace-norm regularization terms are tailored for the fusion of multiple features and the exploitation of the class relationships, respectively. In addition, we impose two additional constraints $\text{tr}(\Psi) = 1$ and $\text{tr}(\Omega) = 1$ to restrict the complexity, as suggested in [43]. In the next section, we introduce an alternating optimization strategy to minimize the above cost function with respect to the network weights $\{\mathbf{W}_l\}_{l=1}^L$, the feature relationship matrix $\Psi$, as well as the class correlation matrix $\Omega$.

### 3.6 Optimization and Analysis

For the optimization problem in Equation (6), two pairs of variables, i.e., $(\mathbf{W}_E, \Psi)$ and $(\mathbf{W}_{L-1}, \Omega)$, are coupled with each other. Here we adopt an alternating optimization approach to iteratively minimize the cost function with respect to $\mathbf{W}_l^m$ $(l = 1, \ldots L, m = 1, \ldots, M)$, $\Psi$ and $\Omega$.

We first consider the minimization problem over the network weight matrix $\mathbf{W}_l^m$ with fixed $\Psi$ and $\Omega$. It is easy to see that the original problem is degenerated to the following unconstrained optimization problem

$$\begin{aligned}
\min_{\mathbf{W}_l^m} \quad & \mathcal{L} + \frac{\lambda_1}{2}\left(\sum_{l=1}^{E}\sum_{m=1}^{M}\|\mathbf{W}_l^m\|_F^2 + \sum_{l=F}^{L-1}\|\mathbf{W}_l\|_F^2\right) \\
& + \frac{\lambda_2}{2}\text{tr}(\mathbf{W}_E\Psi^{-1}\mathbf{W}_E^\top) + \frac{\lambda_3}{2}\text{tr}(\mathbf{W}_{L-1}\Omega^{-1}\mathbf{W}_{L-1}^\top).
\end{aligned} \quad (7)$$

Since all the terms in the above cost function are smooth, the gradient can be easily evaluated. Let $\mathbf{G}_l^m$ be the gradient with respect to $\mathbf{W}_l^m$. We have the following update equation for the weight matrix $\mathbf{W}_l^m$ at the $k$th iteration

$$\mathbf{W}_l^m(k) = \mathbf{W}_l^m(k-1) - \eta\mathbf{G}_l^m(k), \quad (8)$$

where $\eta$ is the step length of the gradient descent.

We then introduce the solution for minimizing the cost function over $\Psi$ with other variables being fixed. The problem in Equation (6) can be rewritten as

$$\begin{aligned}
\min_{\Psi} \quad & \text{tr}(\mathbf{W}_E\Psi^{-1}\mathbf{W}_E^\top), \\
\text{s.t.} \quad & \Psi \succeq 0 \quad \text{tr}(\Psi) = 1.
\end{aligned} \quad (9)$$

Before giving the analytical solution of $\Psi$, we provide a brief discussion on the connection of $\mathbf{W}$ and $\Psi$, which explains the capability of this regularization term in a more rigorous way. We first rewrite the above equation to

$$\begin{aligned}
\min_{\Psi} \quad & \text{tr}(\Psi^{-1}\mathbf{W}_E^\top\mathbf{W}_E), \\
\text{s.t.} \quad & \Psi \succeq 0 \quad \text{tr}(\Psi) = 1.
\end{aligned} \quad (10)$$

Denote $\mathbf{U} = \mathbf{W}^\top\mathbf{W}$, since $\Psi$ is a symmetric matrix and $\text{tr}(\Psi) = 1$ we have

$$\begin{aligned}
\text{tr}(\Psi^{-1}\mathbf{U}) &= \text{tr}(\Psi^{-1}\mathbf{U})\text{tr}(\Psi), \\
&= \left\|\Psi^{-\frac{1}{2}}\mathbf{U}^{\frac{1}{2}}\right\|_F^2 \left\|\Psi^{\frac{1}{2}}\right\|_F^2 \\
&\geq \left\|\Psi^{-\frac{1}{2}}\mathbf{U}^{\frac{1}{2}}\Psi^{\frac{1}{2}}\right\|_F^2 \\
&= (\text{tr}(\mathbf{U}^{\frac{1}{2}}))^2.
\end{aligned} \quad (11)$$

Adopting the Cauchy-Schwarz inequality, $\text{tr}(\Psi^{-1}\mathbf{U})$ attains minimum $(\text{tr}(\mathbf{U}^{\frac{1}{2}}))^2$ if and only if $\Psi^{-\frac{1}{2}}\mathbf{U}^{\frac{1}{2}} = a\Psi^{\frac{1}{2}}$. Therefore, $\Psi$ is determined by matrix $\mathbf{U}$, which defines the relationships among multiple features.

We now provide the analytical solution of $\Psi$ as

$$\Psi = \frac{(\mathbf{W}_E^\top\mathbf{W}_E)^{\frac{1}{2}}}{\text{tr}((\mathbf{W}_E^\top\mathbf{W}_E)^{\frac{1}{2}})}. \quad (12)$$

Similarly, we can derive the optimal solution for $\Omega$ as

$$\Omega = \frac{(\mathbf{W}_{L-1}^\top\mathbf{W}_{L-1})^{\frac{1}{2}}}{\text{tr}((\mathbf{W}_{L-1}^\top\mathbf{W}_{L-1})^{\frac{1}{2}})}. \quad (13)$$

Note that Zhang et al. adopted a similar solution to identify task correlations for a linear kernel based regression and classification problem [43]. However, our method integrates more complex structure regularizations in a neural network architecture, where both the feature and the class relationships are exploited for a completely different application.

In summary, we first estimate the feature and class relationships using the weights in the neural network. The relationship matrices are then utilized in turn to refine the network weights to improve the classification performance. Due to the existence of analytical solutions, we are able to learn the relationship matrices $\Psi$ and $\Omega$ in an efficient way. Finally, the training procedure of the proposed rDNN is summarized in Algorithm 1. In each epoch, we need to compute the gradient matrix $\mathbf{G}_l^m$ for updating $\mathbf{W}_l^m$, and then update the matrices $\Omega$ and $\Psi$. The complexity of calculating the trace norms is the same as that of the $\ell_2$ norm. The update of $\Omega$ and $\Psi$ requires cubic-complexity operations with respect to the number of features $M$ and the number of video classes $C$. In practical large scale settings, the values of $M$ and $C$ are often significantly smaller than the number of training samples. Therefore, the training cost of the proposed rDNN is very similar to that of a standard DNN. Our empirical study further confirms the efficiency of our method, as will be discussed later.

---

**Algorithm 1.** Training Procedure of rDNN

**Require:** $\mathbf{x}_n^m$: the representation of the $m$th feature for the $n$th video sample;
$\mathbf{y}_n$: the semantic label of the $n$th video sample;
1: Initialize $\mathbf{W}_l^m$ randomly, $\Psi = \frac{1}{M}\mathbf{I}_M$ and $\Omega = \frac{1}{C}\mathbf{I}_C$, where $\mathbf{I}_M$ and $\mathbf{I}_C$ are identity matrices;
2: **for** $iteration = 1$ to $K$ **do**
3: 　Back propagate the prediction error from layer $L$ to layer 1 by evaluating the gradient $\mathbf{G}_l^m$, and update the weight matrix $\mathbf{W}_l^m$ for each layer and each feature as:

$$\mathbf{W}_l^m(k) = \mathbf{W}_l^m(k-1) - \eta \mathbf{G}_l^m(k);$$

4: 　Update the feature relationship matrix $\Psi$ according to Equation (12):

$$\Psi = \frac{(\mathbf{W}_E^\top \mathbf{W}_E)^{\frac{1}{2}}}{\operatorname{tr}((\mathbf{W}_E^\top \mathbf{W}_E)^{\frac{1}{2}})};$$

5: 　Update the class relationship matrix $\Psi$ according to Equation (13):

$$\Omega = \frac{(\mathbf{W}_{L-1}^\top \mathbf{W}_{L-1})^{\frac{1}{2}}}{\operatorname{tr}((\mathbf{W}_{L-1}^\top \mathbf{W}_{L-1})^{\frac{1}{2}})}.$$

6: **end for**

---

## 4 EXPERIMENTS

### 4.1 Experimental Setup

#### 4.1.1 Dataset and Evaluation

We adopt three challenging datasets to evaluate the rDNN, as described in the following.

*Hollywood2* [13]. The Hollywood2 dataset is well-known in the area of human action recognition in videos. Collected from 69 Hollywood movies, it contains 1,707 short video clips annotated according to 12 classes: answering phone, driving car, eating, fighting, getting out of car, hand shaking, hugging, kissing, running, sitting down, sitting up and standing up. Following [13], the dataset is split into a training set with 823 videos and a test set with 884 videos.

*Columbia Consumer Videos (CCV)* [58]. The CCV dataset is a popular benchmark on Internet consumer video categorization. It contains 9,317 videos collected from YouTube with annotations of 20 semantic categories, including objects (e.g., "cats"), scenes (e.g., "playground"), and events (e.g., "parade"). Since many categories are events, it requires a joint use of multiple feature clues like visual and audio representations to perform better categorization. The dataset is evenly split into a training set and a test set.

*Fudan-Columbia Video Dataset*. Since both the Hollywood2 and the CCV datasets are small in terms of the number of annotated classes and the number of videos, to substantially evaluate our rDNN, we collect and release a new benchmark, named FCVID.[1] This dataset contains 91,223 Internet videos annotated manually according to 239 categories, covering a wide range of topics like social events (e.g., "tailgate party"), procedural events (e.g., "making cake"), objects (e.g., "panda"), scenes (e.g., "beach"), etc. We divide the dataset evenly with 45,611 videos for training and 45,612 videos for testing. To the best of our knowledge, FCVID is one of the largest datasets for video categorization with accurate manual annotations. Due to space constraint, please refer to the supplementary material, which can be found on the Computer Society Digital Library at http://doi.ieeecomputersociety.org/10.1109/TPAMI.2017.2670560 for more information of the dataset, including details on the collection and annotation process, statistics, a category hierarchy, as well as other related released resources (e.g., all the computed features used in this work).

For all the three datasets, we adopt average precision (AP) to measure the performance of each category and report mean AP (mAP) as the overall results of all the categories. The standard training and testing splits are adopted with no separate validation sets.

#### 4.1.2 Video Features

As aforementioned, we consider both deeply learned features and hand-crafted features in this work.

*Static CNN Features*. Recently, CNN has exhibited top-notch performance in various visual categorization tasks, particularly in the image domain [59]. We adopt a CNN model pre-trained on the ImageNet 2012 Challenge data, which consists of 1.2 million images and 1,000 concept categories. For a given video frame, we extract a 4,096-d feature representation (CNN-$fc_7$), which is the output of the 7th fully connected layer as suggested in [60]. Finally, the frame-level features are averaged to generate a video-level representation.

*Motion Trajectory Features [2]*. The dense trajectory features [2] have been popular for several years, which have exhibited strong performance on various video categorization datasets. Densely sampled local frame patches are first tracked over time to generate the dense trajectories. For each trajectory, four descriptors are computed based on local motion pattern and the appearance around the trajectory, including a 30-d trajectory shape descriptor, a 96-d histogram of oriented gradients (HOG) descriptor, a 108-d histogram of optical flow (HOF) descriptor, and a 108-d motion boundary histogram (MBH) descriptor. Finally, each type of descriptor is quantized into a 4,000-d bag-of-words representation, following the settings of [2].

---

1. Available at: http://bigvid.fudan.edu.cn/FCVID/

*Audio Features.* The audio soundtracks contain very useful clues that can help categorize some video semantics. Two types of video features are considered in this work. The first one is the popular MFCCs (Mel-Frequency Cepstral Coefficients), which are computed over every 32 ms time-window with 50 percent overlap and then quantized into a bag-of-words representation. The second one is called Spectrogram SIFT (sgSIFT) [61], where we transform the 1-d soundtrack of a video into a 2-D image based on the constant-Q spectrogram. Standard SIFT descriptors are extracted from this spectrogram and quantized into a bag-of-words representation.

### 4.1.3 Alternative Approaches for Comparison

To verify the effectiveness of our rDNN, we compare with the following approaches:

1) *DNN*. The same structure with the rDNN using the same 0.5 dropout ratio, but our proposed regularization term is not imposed.
2) *Early Fusion with Neural Networks (NN-EF)*. All the features are concatenated into a long vector and then used as the input to train a neural network for video categorization.
3) *Late Fusion with Neural Networks (NN-LF)*. A neural network is trained using each feature representation independently. The outputs of all the networks are fused to obtain the final categorization results.
4) *Early Fusion with SVM (SVM-EF)*. The popular $\chi^2$ kernel SVM is adopted and the features are combined on the kernel level before classification.
5) *Late Fusion with SVM (SVM-LF)*. An SVM classifier is trained for each feature and prediction results are then combined.
6) *Multiple Kernel Learning (SVM-MKL)*. We perform feature fusion with the $\ell_p$-Norm MKL [62] by fixing $p = 2$. MKL is able to learn dynamic fusion weights. For the above EF/LF approaches 1–4, we adopt equal fusion weights.
7) *Multimodal Deep Boltzmann Machines (M-DBM)*. M-DBM is a fusion approach proposed in [31], where multiple features are used as the inputs of the Deep Boltzmann Machines.
8) *Discriminative Model Fusion (DMF)*. DMF [63] is one of the earliest approaches for exploiting the inter-class relationships. It simply uses the outputs of an initial classifier, e.g., a DNN in our case, as the features to train an SVM model as the second level classifier to generate the final prediction. The second level SVM is expected to be able to learn and use the class relationships.
9) *Domain Adaptive Semantic Diffusion (DASD)*. DASD [37] uses a graph diffusion formulation to utilize the inter-class relationships for visual categorization. Similar to DMF, the prediction outputs of a DNN (without the regularizations) are used as the inputs of the DASD in a post-processing refinement step. The approach requires inputs of pre-computed class correlations, which can be estimated based on statistics of label co-occurrences in the training data. Notice that the pre-computed class correlations are not needed by our rDNN, which can automatically learn the relationships.

Among the alternative approaches, 2–7 focus on feature fusion, while the last two focus on the use of the class relationships. All the neural networks based experiments are conducted on a single NVIDIA Telsa K20 GPU.

### 4.2 Results and Discussion

We now report and discuss experimental results. In order to understand the contributions of only exploiting the feature and the class relationships, we first test the performance of the rDNN by disabling the regularizations on the output layer and the fusion layer, respectively. This also ensures to make fair comparisons with the alternative approaches. After that, we enable regularizations on both layers and report results of the entire rDNN framework. With this setting, we analyze the effect of the number of training samples, and compare with recent state-of-the-art results. Last, we discuss the computational efficiency of rDNN.

Throughout the experiments, we use 4 layers of neurons in the rDNN. All the features are used as the input of the first layer, which are then transformed using a hidden layer with 256 neurons for each type of feature separately. The transformed features are further fused with a fusion layer containing 256 neurons, and the fused feature is finally converted to classification scores through the last layer. Note that 4 layers are empirically found to be suitable. Using more layers in rDNN may improve the results but would probably require more training data.

For the key parameters, we set the learning rate of the neural networks to 0.7, fix $\lambda_1$ to a small value of $3e^{-5}$ in order to prevent overfitting, and set $\lambda_2$ and $\lambda_3$ to 5e-5 for Hollywood and CCV, and 3e-5 for FCVID. We adopt the mini batch gradient descent with the batch size being 70 for network training. The training will stop if it reaches the maximal epochs or the training error stops to decrease in the last 10 epochs (with difference less than 1e-5).

### 4.2.1 Exploiting Feature Relationships

We first report results by only using the fusion layer regularization in our rDNN, namely rDNN-Feature Regularization (rDNN-F). Table 1 shows the results of the individual features, our rDNN-F, and the alternative feature fusion methods. Among the static CNN, motion and audio features, motion is significant better than the other two on Hollywood2 but is slightly worse than the CNN feature on CCV and FCVID. This is due to the fact that many classes in CCV and FCVID (e.g., "baseball" and "desert") can be recognized by viewing just one or a few discrete frames, but categorizing the Hollywood2 human actions normally requires a sequence of frames with detailed motion clues. In addition, the overall performance on CCV is slightly lower than that on the much larger FCVID. This is because CCV has some highly correlated categories (see Fig. 4) that are

TABLE 1
Performance Comparison (mAP) Using Individual and Multiple Features with Various Fusion Methods

|  | Hollywood2 | CCV | FCVID |
| --- | --- | --- | --- |
| Static CNN | 40.1% | 66.1% | 63.8% |
| Motion | 62.4% | 60.8% | 62.8% |
| Audio | 22.7% | 25.9% | 26.1% |
| DNN | 64.2% | 71.6% | 72.1% |
| NN-EF | 63.5% | 70.2% | 74.7% |
| NN-LF | 60.2% | 69.9% | 73.8% |
| SVM-EF | 64.1% | 71.7% | 75.0% |
| SVM-LF | 62.7% | 69.1% | 72.1% |
| SVM-MKL [62] | 63.8% | 71.3% | 75.2% |
| M-DBM [31] | 63.9% | 71.1% | 74.4% |
| rDNN-F | 65.9% | 72.9% | 75.4% |

"rDNN-F" indicates our rDNN focusing only on the exploitation of the feature relationships.

TABLE 2
Performance Comparison (mAP) with DMF and DASD, which Focus on the Use of the Class Relationships

|  | Hollywood2 | CCV | FCVID |
| --- | --- | --- | --- |
| DNN | 64.2% | 71.6% | 72.1% |
| DMF [63] | 61.8% | 71.1% | 72.5% |
| DASD [37] | 64.4% | 71.7% | 72.8% |
| rDNN-C | 65.1% | 72.1% | 74.4% |
| rDNN-C prior | 65.8% | 72.5% | 75.0% |

"DNN" is a baseline without imposing our proposed regularization term and "rDNN-C" indicates our rDNN utilizing only the class relationships.

very difficult to be separated. While FCVID also contains similar confusing categories, the percentage of such "difficult" cases is lower as it also has more "easy" categories, and therefore the overall performance is higher.

For the fusion of the three types of features, our rDNN-F achieves the best performance with consistent gains over all the compared methods. Note that, like the "DNN" baseline, the M-DBM approach also utilizes a neural network for feature fusion, but in a *free* manner without explicitly enforcing the use of the feature relationships. These results clearly verifies the effectiveness of imposing the proposed fusion regularization method. Notice that, since the Hollywood2 and the CCV datasets have been widely used, an absolute mAP gain of 2 percent is generally considered as a significant improvement.

Among the alternative approaches, early fusion methods tend to produce better results than late fusion. This is consistent with the observations of several recent works, where early fusion is more popularly adopted [3]. The MKL is even slightly worse than early fusion on Hollywood2 and CCV, indicating that the learned weights do not generalize well to testing data. In addition, for the contribution of the audio feature in the fusion experiments, we observe clearly improvement for the classes with strong audio clues, such as "answering phone". On the contrary, for classes like "sitting down", audio features may slightly degrade the result.

#### 4.2.2 Exploiting Class Relationships

Next, we report results of rDNN using only the class relationships, namely rDNN-C. We compare with the DNN baseline with no regularization, DMF and DASD. Results are given in Table 2. rDNN-C outperforms the DNN baseline and the two alternative approaches. Both DMF and DASD use the outputs of the DNN baseline as inputs for context-based refinement. These results corroborate the effectiveness of the class relationship regularization.

Note that, like many previous methods exploring class relationships, the DASD requires pre-computed class relationships as the input, which are estimated based on the label co-occurrences in the training data. This might be the reason that it performs worse than the rDNN-C as the latter automatically learns the commonalities shared among the categories. The learning process can identify not only the categories that co-occur, but also those sharing visual or auditory commonalities but rarely appear together. To shown in Table 2 (the last row rDNN-C *prior*), we observe further improvements, which are however not very significant. Therefore, we conclude that the automatically identified visual/auditory commonalities are effective. In addition, to further verify that the gain of our approach is really from using the learned class relationships, not from our different optimization strategy as compared with the baselines, we simply fix $\Omega$ to be an identity matrix in the optimization process (i.e., all the classes are treated independently). Under this setting, the performance drops 1.6, 0.8 and 3.0 percent on Hollywood2, CCV and FCVID respectively.

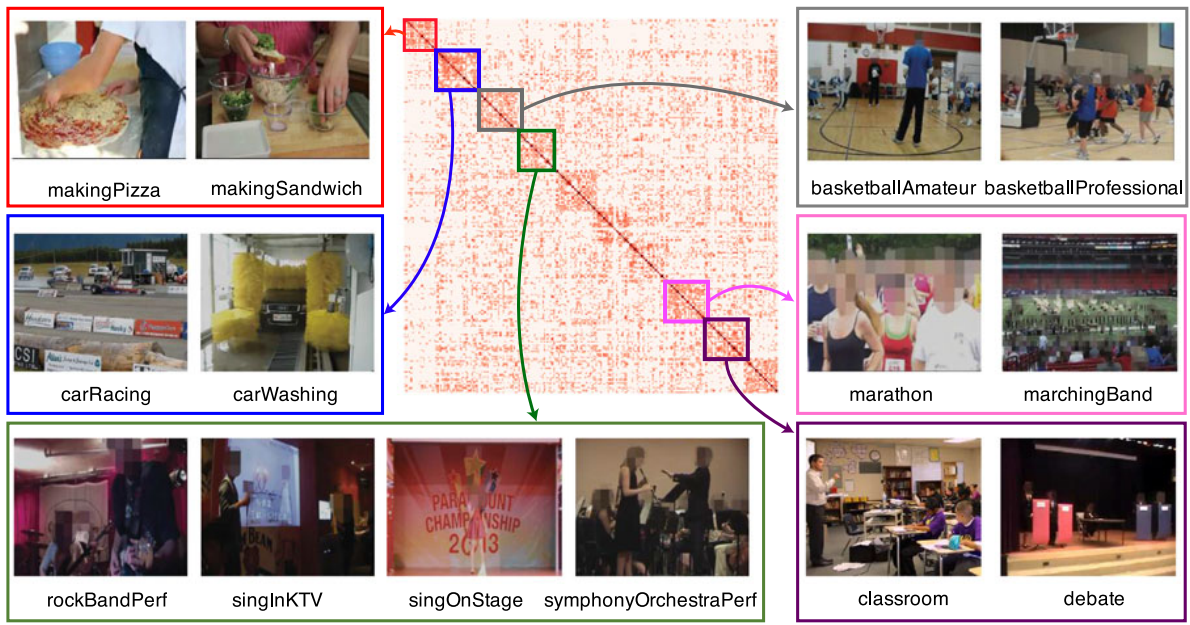

Fig. 3. The learned class relationship matrix $\Omega$ on FCVID and example frames of a few category groups. Many of the found groups contain categories that share visual/auditory commonalities but do not necessarily co-occur.

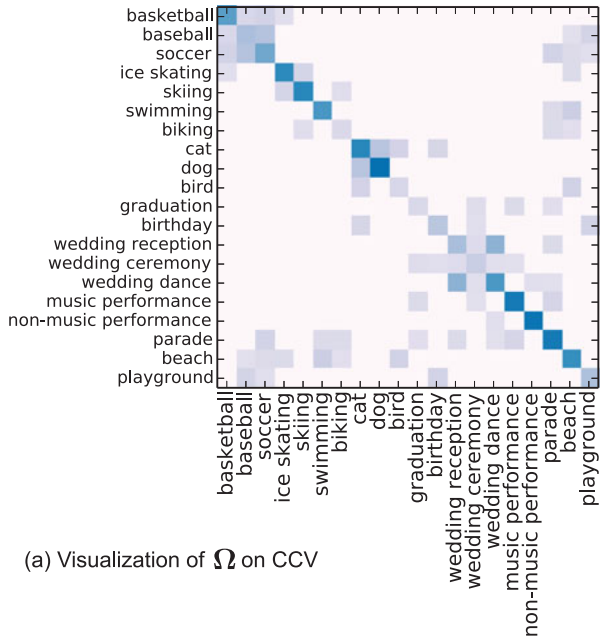

(a) Visualization of $\Omega$ on CCV

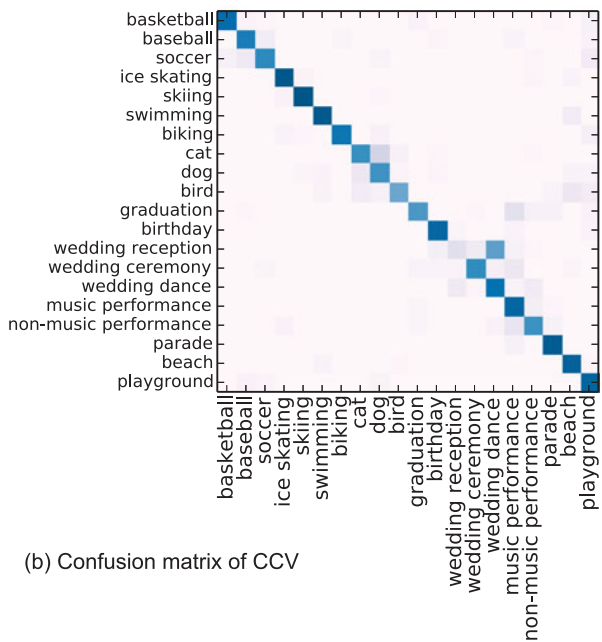

(b) Confusion matrix of CCV

Fig. 4. (a) The learned class relationship matrix $\Omega$ on CCV. (b) Confusion matrix on CCV.

TABLE 3
Performance of the Entire Framework (the Last Row) Using Both Kinds of Relationships, in Comparison with Single-Relationship Results and the Basic Deep Networks with Various Numbers of Hidden Layers

|  | Hollywood2 | CCV | FCVID |
| --- | --- | --- | --- |
| DNN | 64.2% | 71.6% | 72.1% |
| DNN-6 layer | 60.1% | 68.1% | 67.3% |
| DNN-8 layer | 56.2% | 62.3% | 62.7% |
| rDNN-F | 65.9% | 72.9% | 75.4% |
| rDNN-C | 65.1% | 72.1% | 74.4% |
| rDNN | 66.9% | 73.5% | 76.0% |

### 4.2.3 Exploiting Both Kinds of Relationships

Finally, we discuss the results of the entire rDNN framework, using both the feature and the class relationships. Table 3 presents the results of the overall framework. Overall, substantial performance gains are attained from the proposed approach. Using regularizations on both kinds of relationships leads to clearly higher performance than imposing the regularization on a single type of relationship.

Compared with DNN structures that only adopt dropout to improve generalization, rDNN achieves better performance on all the datasets. We also deepened the network structures with six and eight layers in order to learn the hidden relationships (indicated by "DNN-6 layer" and "DNN-8 layer" in the table), but the results are significantly worse. This is because more parameters are added with the additional layers, which will easily lead to over-fitting especially when training with limited samples.

In addition, comparing the results across the three datasets, the improvement from exploiting the class relationships is more significant on FCVID. This is because FCVID contains a much larger number of classes that share commonalities helpful for categorization. Fig. 4 further visualizes the confusion matrix of rDNN on the CCV dataset.

### 4.2.4 Training with Limited Samples

Regularization techniques could usually help improve the results when training with limited samples. To better evaluate the effectiveness of the regularizers, we plot the performance with different numbers of training samples in Fig. 5. We observe that the performance gain of rDNN is more significant when the number of training samples is small (except the case of 10 training samples on FCVID, which are too few to distinguish the 239 categories). Under all the settings, the rDNN requires less training data to achieve comparable results to the non-regularized version.

verify this, we visualize some found category groups in Fig. 3. As discussed in Section 3, values in the matrix $\Omega$ can indicate the learned relationships among the categories. Hence, we apply the spectral clustering algorithm on $\Omega$ to group the categories and provide examples of several classes having high similarities. We see that many categories are grouped together because they share certain commonalities (e.g., "marathon" and "marchingBand"), not due to high frequencies of co-occurrence. In addition, we further visualize the learned matrix $\Omega$ on the smaller dataset CCV in Fig. 4a. The learned correlated categories may be due to either the shared objects, scenes or audio clues.

It is interesting to notice that, once prior knowledge of the relations among multiple categories is available, it can be leveraged to initialize the relationship matrix $\Omega$. For example, if a category $i$ is known to be more similar to $j$ than to $k$, we could simply set $\Omega_{ij} > \Omega_{ik}$ in the initialization stage. As

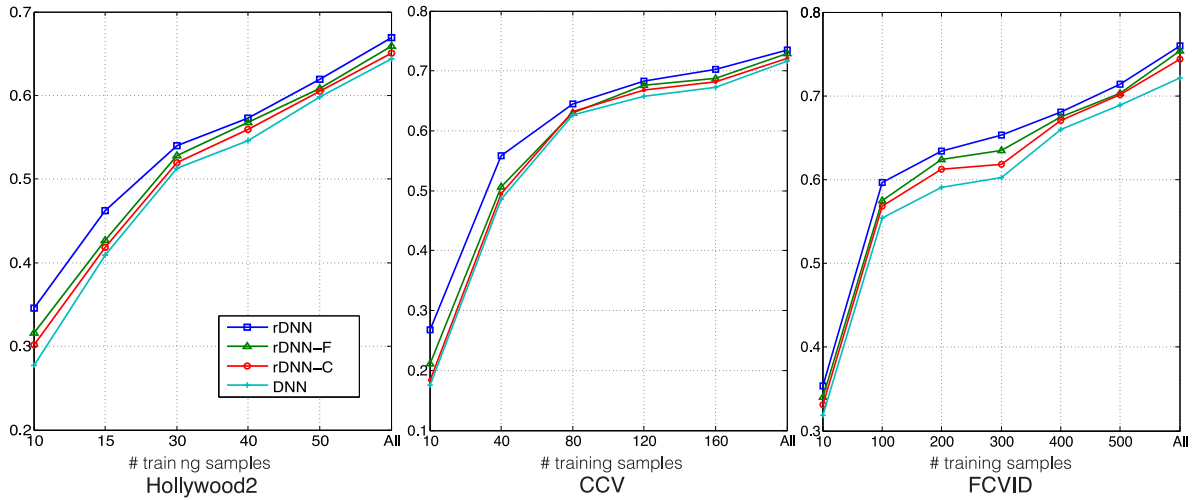

Fig. 5. Performance of different numbers of training samples. We plot the results of the DNN baseline without regularization, rDNN-F, rDNN-C and the rDNN exploiting both types of relationships. The best mAP on the three datasets (the rDNN approach using all the training samples) are 66.9, 73.5 and 76.0 percent respectively.

#### 4.2.5 Comparison with State of the Arts

We compare rDNN with several recent approaches in Table 4. On Hollywood2, our proposed method achieves a competitive mAP of 66.9 percent, outperforming many of the compared approaches [2], [11], [64], [65], [66], except a few recent results [67], [68], [69]. Most of these approaches are based on the popular dense trajectory features and the SVM classification with the simple early fusion method. Note that some of them like Wang et al. [2] and Lan et al. [67] encoded the features using the Fisher vector [70], which has been shown to be more effective than the classical bag-of-words representation used in our approach. The approach by Lan et al. [67] extends upon the dense trajectories with a feature enhancement method called multi-skip feature stacking, while Hoai et al. [68] and Fernando et al. [69] explored prediction score distribution and temporal information respectively. Since the focus of these works is different, further performance gain may be achieved by combining them with rDNN.

On the CCV dataset, we obtain to-date the best performance with an mAP of 73.5 percent. Most recent works on CCV focused on the joint use of multiple audio-visual features. Xu et al. [72] and Ye et al. [27] extended late fusion with specially designed strategies to remove the noise of individually trained classifiers. Jhuo et al. adopted a joint audio-visual codebook to exploit feature relationships for categorization [29].

TABLE 4
Comparison with State of the Arts

| Hollywood2 | mAP | CCV | mAP |
| --- | --- | --- | --- |
| Jain et al. [64] | 62.5% | Kim et al. [71] | 56.5% |
| Wang et al. [2] | 64.3% | Xu et al. [72] | 60.3% |
| Zhang et al. [65] | 50.9% | Ye et al. [27] | 64.0% |
| Ni et al. [66] | 61.0% | Jhuo et al. [29] | 64.0% |
| Wu et al. [11] | 65.7% | Ma et al. [73] | 63.4% |
| Lan et al. [67] | 68.0% | Liu et al. [74] | 68.2% |
| Hoai et al. [68] | 73.6% | Wu et al. [11] | 70.6% |
| Fernando et al. [69] | 73.7% | - | - |
| **rDNN** | **66.9%** | **rDNN** | **73.5%** |

#### 4.2.6 Computational Efficiency

We discuss the computational efficiency of rDNN using the Hollywood2 dataset. The average training time of each epoch for NN-EF, NN-LF and rDNN are $1.540 \pm 0.02$, $1.552 \pm 0.05$ and $1.276 \pm 0.10$, respectively, using the same GPU-based implementation. rDNN is more efficient than NN-EF and NN-LF as it contains less parameters to be learned. Specifically, compared with the NN-EF, rDNN processes the features separately in the first two layers and thus avoids the parameters needed for interacting among them. The NN-LF requires the training of separate networks, which is also more expensive. Note that the M-DBM method is not compared because it requires much more time to pre-train the network for weight initialization. For all the methods, normally a few hundreds of epochs are needed to finish the training (several minutes in total). After training, all the neural network methods are very fast in testing.

### 5 CONCLUSION

We have proposed a novel rDNN approach to exploit both feature and class relationships in video categorization. By imposing trace-norm based regularizations on the specially tailored fusion layer and output layer, our rDNN can learn a fused representation of multiple feature inputs and utilize the commonalities shared among the semantic classes for improved categorization performance. Extensive experiments of action and event recognition on popular benchmarks have shown that rDNN consistently outperforms several alternative approaches. Our rDNN is also efficient in both model training and testing, which is very important for large scale applications. In addition, we have introduced a new dataset, FCVID, for large scale video categorization. We believe that FCVID is helpful for stimulating research not only on video categorization, but also on other related problems.

The current framework supports the use of any pre-computed features. One interesting future work is to exploit the joint learning of feature representations and classification models. For instance, the adopted CNN feature is computed based on off-the-shelf models. It would be probably helpful if the feature extraction part could be further tuned simultaneously with the regularized classification network.

## ACKNOWLEDGMENTS

This work was supported in part by two grants from NSF China (#61622204, #61572138) and a grant from STCSM, Shanghai, China (#16JC1420401).

## REFERENCES

[1] A. Krizhevsky, I. Sutskever, and G. E. Hinton, "Imagenet classification with deep convolutional neural networks," in *Proc. Advances Neural Inf. Process. Syst.*, 2012, pp. 1106–1114.
[2] H. Wang and C. Schmid, "Action recognition with improved trajectories," in *Proc. IEEE Int. Conf. Comput. Vis.*, 2013, pp. 3551–3558.
[3] R. Aly, et al., "The AXES submissions at TrecVid 2013," in *Proc. NIST TRECVID Workshop*, 2013.
[4] Z.-Z. Lan, et al., "CMU-Informedia@TRECVID 2013 multimedia event detection," in *Proc. NIST TRECVID Workshop*, 2013.
[5] A. Karpathy, G. Toderici, S. Shetty, T. Leung, R. Sukthankar, and L. Fei-Fei, "Large-scale video classification with convolutional neural networks," in *Proc. IEEE Conf. Comput. Vis. Pattern Recog.*, 2014, pp. 1725–1732.
[6] K. Simonyan and A. Zisserman, "Two-stream convolutional networks for action recognition in videos," in *Proc. Advances Neural Inf. Process. Syst.*, 2014, pp. 568–576.
[7] C. Feichtenhofer, A. Pinz, and A. Zisserman, "Convolutional two-stream network fusion for video action recognition," in *Proc. IEEE Conf. Comput. Vis. Pattern Recog.*, 2016, pp. 1933–1941.
[8] G. Ye, Y. Li, H. Xu, D. Liu, and S.-F. Chang, "EventNet: A large scale structured concept library for complex event detection in video," in *Proc. 23rd ACM Int. Conf. Multimedia*, 2015, pp. 471–480.
[9] F. C. Heilbron, V. Escorcia, B. Ghanem, and J. C. Niebles, "ActivityNet: A large-scale video benchmark for human activity understanding," in *Proc. IEEE Conf. Comput. Vis. Pattern Recog.*, 2015, pp. 961–970.
[10] M. Andriluka, L. Pishchulin, P. Gehler, and B. Schiele, "2D human pose estimation: New benchmark and state of the art analysis," in *Proc. IEEE Conf. Comput. Vis. Pattern Recog.*, 2014, pp. 3686–3693.
[11] Z. Wu, Y.-G. Jiang, J. Wang, J. Pu, and X. Xue, "Exploring inter-feature and inter-class relationships with deep neural networks for video classification," in *Proc. 22 ACM Int. Conf. Multimedia*, 2014, pp. 167–176.
[12] H. Jhuang, T. Serre, L. Wolf, and T. Poggio, "A biologically inspired system for action recognition," in *Proc. IEEE Int. Conf. Comput. Vis.*, 2007, pp. 1–8.
[13] I. Laptev, M. Marszalek, C. Schmid, and B. Rozenfeld, "Learning realistic human actions from movies," in *Proc. IEEE Conf. Comput. Vis. Pattern Recog.*, 2008, pp. 1–8.
[14] C. V. Cotton and D. P. W. Ellis, "Subband autocorrelation features for video soundtrack classification," in *Proc. IEEE Int. Conf. Acoust. Speech Signal Process.*, 2013, pp. 8663–8666.
[15] S. Ji, W. Xu, M. Yang, and K. Yu, "3D convolutional neural networks for human action recognition," in *Proc. Int. Conf. Mach. Learn.*, 2010, pp. 495–502.
[16] N. Srivastava, E. Mansimov, and R. Salakhutdinov, "Unsupervised learning of video representations using LSTMs," *ICML*, pp. 843–852, 2015.
[17] J. Y.-H. Ng, M. Hausknecht, S. Vijayanarasimhan, O. Vinyals, R. Monga, and G. Toderici, "Beyond short snippets: Deep networks for video classification," in *Proc. IEEE Conf. Comput. Vis. Pattern Recog.*, 2015, pp. 4694–4702.
[18] Z. Wu, X. Wang, Y.-G. Jiang, H. Ye, and X. Xue, "Modeling spatial-temporal clues in a hybrid deep learning framework for video classification," in *Proc. 23rd ACM Int. Conf. Multimedia*, 2015, pp. 461–470.
[19] S. Maji, A. C. Berg, and J. Malik, "Classification using intersection kernel support vector machines is efficient," in *Proc. IEEE Conf. Comput. Vis. Pattern Recog.*, 2008, pp. 1–8.
[20] Y.-G. Jiang, "Super: Towards real-time event recognition in internet videos," in *Proc. 2nd ACM Int. Conf. Multimedia Retrieval*, 2012, Art. no. 7.
[21] Y. Zou, X. Jin, Y. Li, Z. Guo, E. Wang, and B. Xiao, "Mariana: Tencent deep learning platform and its applications," *Proc. VLDB Endowment*, vol. 7, pp. 1772–1777, 2014.
[22] O. Yadan, K. Adams, Y. Taigman, and M. Ranzato, "Multi-GPU training of convnets," *arXiv preprint arXiv:1312.5853*, vol. 9, 2013.
[23] F. R. Bach, G. R. Lanckriet, and M. I. Jordan, "Multiple kernel learning, conic duality, and the SMO algorithm," in *Proc. 21st Int. Conf. Mach. Learn.*, 2004, Art. no. 6.
[24] L. Cao, J. Luo, F. Liang, and T. S. Huang, "Heterogeneous feature machines for visual recognition," in *Proc. IEEE Int. Conf. Comput. Vis.*, 2009, pp. 1095–1102.
[25] P. Natarajan, et al., "Multimodal feature fusion for robust event detection in web videos," in *Proc. IEEE Conf. Comput. Vis. Pattern Recog.*, 2012, pp. 1298–1305.
[26] A. Vedaldi, V. Gulshan, M. Varma, and A. Zisserman, "Multiple kernels for object detection," in *Proc. IEEE Int. Conf. Comput. Vis.*, 2009, pp. 606–613.
[27] G. Ye, D. Liu, I.-H. Jhuo, and S.-F. Chang, "Robust late fusion with rank minimization," in *Proc. IEEE Conf. Comput. Vis. Pattern Recog.*, 2012, pp. 3021–3028.
[28] W. Jiang, C. Cotton, S.-F. Chang, D. Ellis, and A. Loui, "Short-term audio-visual atoms for generic video concept classification," in *Proc. 17th ACM Int. Conf. Multimedia*, 2009, pp. 5–14.
[29] I.-H. Jhuo, et al., "Discovering joint audio-visual codewords for video event detection," *Mach. Vis. Appl.*, vol. 25, pp. 33–47, 2014.
[30] J. Ngiam, A. Khosla, M. Kim, J. Nam, H. Lee, and A. Ng, "Multimodal deep learning," in *Proc. Int. Conf. Mach. Learn*, 2011, pp. 689–696.
[31] N. Srivastava and R. Salakhutdinov, "Multimodal learning with deep Boltzmann machines," in *Proc. Advances Neural Inf. Process. Syst.*, 2012, pp. 2231–2239.
[32] K. Sohn, W. Shang, and H. Lee, "Improved multimodal deep learning with variation of information," in *Proc. Advances Neural Inf. Process. Syst.*, 2014, pp. 2141–2149.
[33] A. Torralba, "Contextual priming for object detection," *Int. J. Comput. Vis.*, vol. 53, pp. 169–191, 2003.
[34] S. Bengio, J. Dean, D. Erhan, E. Ie, Q. Le, A. Rabinovich, J. Shlens, and Y. Singer, "Using web co-occurrence statistics for improving image categorization," *arXiv preprint arXiv:1312.5697*, 2013.
[35] A. Rabinovich, A. Vedaldi, C. Galleguillos, E. Wiewiora, and S. Belongie, "Objects in context," in *Proc. IEEE Int. Conf. Comput. Vis.*, 2007, pp. 1–8.
[36] H. R. Naphade and T. S. Huang, "A probabilistic framework for semantic video indexing, filtering, and retrieval," *IEEE Trans. Multimedia*, vol. 3, no. 1, pp. 141–151, Mar. 2001.
[37] Y.-G. Jiang, Q. Dai, J. Wang, C.-W. Ngo, X. Xue, and S.-F. Chang, "Fast semantic diffusion for large-scale context-based image and video annotation," *IEEE Trans. Image Process.*, vol. 21, no. 6, pp. 3080–3091, Jun. 2012.
[38] M.-F. Weng and Y.-Y. Chuang, "Cross-domain multicue fusion for concept-based video indexing," *IEEE Trans. Pattern Anal. Mach. Intell.*, vol. 34, no. 10, pp. 1927–1941, Oct. 2012.
[39] J. Deng, et al., "Large-scale object classification using label relation graphs," in *Proc. 13th Eur. Conf. Comput. Vis.*, 2014, pp. 48–64.
[40] S. M. Assari, A. R. Zamir, and M. Shah, "Video classification using semantic concept co-occurrences," in *Proc. IEEE Conf. Comput. Vis. Pattern Recog.*, 2014, pp. 2529–2536.
[41] T. Mensink, E. Gavves, and C. G. M. Snoek, "COSTA: Co-occurrence statistics for zero-shot classification," in *Proc. IEEE Conf. Comput. Vis. Pattern Recog.*, 2014, pp. 2441–2448.
[42] L. Jacob, F. R. Bach, and J.-P. Vert, "Clustered multi-task learning: A convex formulation," in *Proc. Advances Neural Inf. Process. Syst.*, 2008, pp. 745–752.
[43] Y. Zhang and D.-Y. Yeung, "A convex formulation for learning task relationships in multi-task learning," in *Proc. 26th Conf. Uncertainty Artif. Intell.*, 2010, pp. 733–742.
[44] D. Zhang and D. Shen, "Multi-modal multi-task learning for joint prediction of multiple regression and classification variables in Alzheimer's disease," *Neuroimage*, vol. 59, pp. 895–907, 2012.
[45] J. Zhou, L. Yuan, J. Liu, and J. Ye, "A multi-task learning formulation for predicting disease progression," in *Proc. 17th ACM SIGKDD Int. Conf. Knowl. Discovery Data Mining*, 2011, pp. 814–822.
[46] J. Ghosn and Y. Bengio, "Multi-task learning for stock selection," in *Proc. Advances Neural Inf. Process. Syst.*, 1997, pp. 946–952.
[47] J. Chen, J. Zhou, and J. Ye, "Integrating low-rank and group-sparse structures for robust multi-task learning," in *Proc. 17th ACM SIGKDD Int. Conf. Knowl. Discovery Data Mining*, 2011, pp. 42–50.
[48] Z. Kang, K. Grauman, and F. Sha, "Learning with whom to share in multi-task feature learning," in *Proc. Int. Conf. Mach. Learn.*, 2011, pp. 521–528.

[49] J. Pu, Y.-G. Jiang, J. Wang, and X. Xue, "Multiple task learning using iteratively reweighted least square," in *Proc. 23rd Int. Joint Conf. Artif. Intell.*, 2013, pp. 1607–1613.
[50] R. Caruana, "Multitask learning," *Mach. Learn.*, vol. 28, pp. 41–75, 1997.
[51] T. Ohshiro, D. Angelaki, and G. DeAngelis, "A normalization model of multisensory integration," *Nature Neuroscience*, vol. 14, pp. 775–782, 2011.
[52] B. E. Stein and T. R. Stanford, "Multisensory integration: Current issues from the perspective of the single neuron," *Nature Rev. Neuroscience*, vol. 9, pp. 255–266, 2008.
[53] H. Fei and J. Huan, "Structured feature selection and task relationship inference for multi-task learning," *Knowl. Inf. Syst.*, vol. 35, pp. 345–364, 2013.
[54] G.-J. Qi, X.-S. Hua, Y. Rui, J. Tang, T. Mei, and H.-J. Zhang, "Correlative multi-label video annotation," in *Proc. 15th ACM Int. Conf. Multimedia*, 2007, pp. 17–26.
[55] G. Ye, D. Liu, J. Wang, and S.-F. Chang, "Large-scale video hashing via structure learning," in *Proc. IEEE Int. Conf. Comput. Vis.*, 2013, pp. 2272–2279.
[56] A. Argyriou, T. Evgeniou, and M. Pontil, "Convex multi-task feature learning," *Mach. Learn.*, vol. 73, pp. 243–272, 2008.
[57] J. Liu, S. Ji, and J. Ye, "Multi-task feature learning via efficient l 2, 1-norm minimization," in *Proc. 25th Conf. Uncertainty Artif. Intell.*, 2009, pp. 339–348.
[58] Y.-G. Jiang, G. Ye, S.-F. Chang, D. Ellis, and A. C. Loui, "Consumer video understanding: A benchmark database and an evaluation of human and machine performance," in *Proc. 1st ACM Int. Conf. Multimedia Retrieval*, 2011, Art. no. 29.
[59] C. Szegedy, W. Liu, Y. Jia, P. Sermanet, S. Reed, D. Anguelov, D. Erhan, V. Vanhoucke, and A. Rabinovich, "Going deeper with convolutions," in *Proc. IEEE Conf. Comput. Vis. Pattern Recog.*, 2015, pp. 1–9.
[60] R. Girshick, J. Donahue, T. Darrell, and J. Malik, "Rich feature hierarchies for accurate object detection and semantic segmentation," in *Proc. IEEE Conf. Comput. Vis. Pattern Recog.*, 2014, pp. 580–587.
[61] B. Zhu, W. Li, Z. Wang, and X. Xue, "A novel audio fingerprinting method robust to time scale modification and pitch shifting," in *Proc. 18th ACM Int. Conf. Multimedia*, 2010, pp. 987–990.
[62] M. Kloft, U. Brefeld, S. Sonnenburg, and A. Zien, "Lp-norm multiple kernel learning," *J. Mach. Learn. Res.*, vol. 12, pp. 953–997, 2011.
[63] J. R. Smith, M. Naphade, and A. Natsev, "Multimedia semantic indexing using model vectors," in *Proc. IEEE Int. Conf. Multimedia Expo*, 2003, pp. 445–448.
[64] M. Jain, H. Jégou, and P. Bouthemy, "Better exploiting motion for better action recognition," in *Proc. IEEE Conf. Comput. Vis. Pattern Recog.*, 2013, pp. 2555–2562.
[65] H. Zhang, W. Zhou, C. M. Reardon, and L. E. Parker, "Simplex-based 3D spatio-temporal feature description for action recognition," in *Proc. IEEE Conf. Comput. Vis. Pattern Recog.*, 2014, pp. 2067–2074.
[66] B. Ni, T. Li, and P. Moulin, "Beta process multiple kernel learning," in *Proc. IEEE Conf. Comput. Vis. Pattern Recog.*, 2014, pp. 963–970.
[67] Z. Lan, M. Lin, X. Li, A. G. Hauptmann, and B. Raj, "Beyond Gaussian pyramid: Multi-skip feature stacking for action recognition," in *Proc. IEEE Conf. Comput. Vis. Pattern Recog.*, 2015, pp. 204–212.
[68] M. Hoai and A. Zisserman, "Improving human action recognition using score distribution and ranking," in *Proc. Asian Conf. Comput. Vis.*, 2014, pp. 3–20.
[69] B. Fernando, E. Gavves, J. M. Oramas, A. Ghodrati, and T. Tuytelaars, "Modeling video evolution for action recognition," in *Proc. IEEE Conf. Comput. Vis. Pattern Recog.*, 2015, pp. 5378–5387.
[70] J. Sánchez, F. Perronnin, T. Mensink, and J. Verbeek, "Image classification with the Fisher vector: Theory and practice," *Int. J. Comput. Vis.*, vol. 105, pp. 222–245, 2013.
[71] I. Kim, S. Oh, B. Byun, A. A. Perera, and C.-H. Lee, "Explicit performance metric optimization for fusion-based video retrieval," in *Proc. Eur. Conf. Comput. Vis. Workshop*, 2012, pp. 395–405.
[72] Z. Xu, Y. Yang, I. Tsang, N. Sebe, and A. Hauptmann, "Feature weighting via optimal thresholding for video analysis," in *Proc. IEEE Int. Conf. Comput. Vis.*, 2013, pp. 3440–3447.
[73] A. J. Ma and P. C. Yuen, "Reduced analytic dependency modeling: Robust fusion for visual recognition," *Int. J. Comput. Vis.*, vol. 109, pp. 233–251, 2014.
[74] D. Liu, K.-T. Lai, G. Ye, M.-S. Chen, and S.-F. Chang, "Sample-specific late fusion for visual category recognition," in *Proc. IEEE Conf. Comput. Vis. Pattern Recog.*, 2013, pp. 803–810.

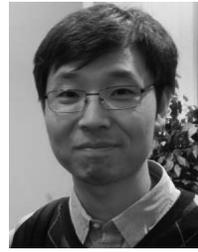

**Yu-Gang Jiang** received the PhD degree in computer science from City University of Hong Kong, Kowloon, Hong Kong, in 2009. During 2008-2011, he was in the Department of Electrical Engineering, Columbia University, New York. He is currently a professor of computer science with Fudan University, Shanghai, China. His research interests include computer vision and multimedia.

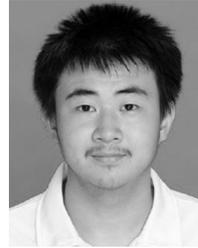

**Zuxuan Wu** received the BEng degree in software engineering from East China Normal University, in 2013 and the MSc degree from Fudan Univeristy, in 2016. He is currently working toward the PhD degree in the Computer Science Department, University of Maryland, College Park. His research interests include computer vision, multimedia retrieval, and deep learning.

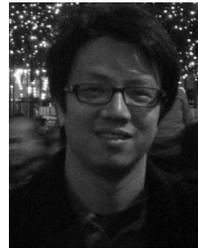

**Jun Wang** received the PhD degree from Columbia University, New York, in 2011. Currently, he is a professor of computer science with East China Normal University, Shanghai, China. His research interests include machine learning, information retrieval, and data mining.

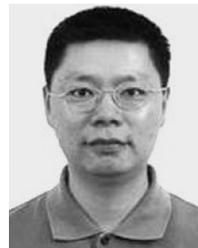

**Xiangyang Xue** received the BS, MS, and PhD degrees in communication engineering from Xidian University, Xi'an, China, in 1989, 1992, and 1995, respectively. He is currently a professor of computer science with Fudan University, Shanghai, China. His research interests include multimedia information processing and machine learning.

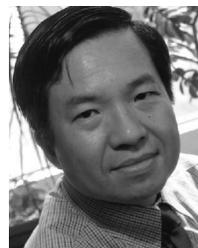

**Shih-Fu Chang** is the Richard Dicker professor, director of the Digital Video and Multimedia Lab, and senior vice dean of Engineering School, Columbia University. His research is focused on multimedia information retrieval, computer vision, machine learning, and signal processing, with the goal to turn unstructured multimedia data into searchable information. He is a fellow of the American Association for the Advancement of Science (AAAS) and the IEEE.

▷ **For more information on this or any other computing topic, please visit our Digital Library at** www.computer.org/publications/dlib.